%% file: main.tex
\documentclass[twoside,leqno,twocolumn]{article}

\usepackage{epstopdf}

\usepackage{amssymb}
\usepackage{graphicx}

\usepackage{float}
\usepackage{amsmath}
\usepackage{graphicx}
\usepackage{graphics}
\usepackage{caption}
\usepackage{subfig}
\usepackage{url}
\usepackage{amsmath}
\usepackage{amssymb}
\usepackage{epstopdf}
\usepackage{wasysym}
\usepackage{multirow}
\usepackage{graphicx}
\usepackage{caption}

\usepackage{lipsum}
\usepackage{array}

\usepackage{subfig}
\usepackage[super,sort&compress]{natbib}
\usepackage{soul}
\usepackage{multirow}
\usepackage{algorithm}
\usepackage{algorithmic}
\usepackage[margin=1in]{geometry}
\newcommand{\Fscore}{\ensuremath{\textit{F-score}}}%
\newcommand{\etal}{\textit{et al.}}

\begin{document}

	\title{A Personalized Federated Learning Algorithm: an Application in Anomaly  Detection}
	\author{Ali Anaissi
		\and Basem Suleiman}

\maketitle
	
	\begin{abstract}
	Federated Learning (FL) has recently emerged as a promising method that employs a distributed learning model structure to overcome data privacy and transmission issues paused by central machine learning models. In FL, datasets collected from different devices or sensors are used to train local models (clients) each of which shares its learning with a centralized model (server). However, this distributed learning approach presents unique learning challenges as the data used at local clients can be non-IID (Independent and Identically Distributed) and statistically diverse which decrease learning accuracy in the central model. In this paper, we overcome this problem by proposing a novel Personalized Conditional FedAvg (PC-FedAvg) which aims to control weights communication and aggregation augmented with   a tailored learning algorithm to personalize the resulting models at each client. Our experimental validation on two datasets showed that our PC-FedAvg precisely constructed generalized clients' models and thus achieved higher accuracy compared to other  state-of-the-art methods.	
	\end{abstract}
	
	\input{introduction.tex}

	\input{relatedwork.tex}
	
	
	\input{method.tex}

	\input{experiments.tex}

\input{conclusion.tex}
	
	\bibliographystyle{siamplain}
	\scriptsize
	\begingroup
	\setlength{\bibsep}{0pt}
	 
	\bibliography{myBib}
	\endgroup

\end{document}

%% file: introduction.tex
\section{Introduction}
\label{s:introduction}

The emerging  Federated Learning (FL) concept was initially proposed by Google for improving security and preventing data leakages in distributed environments \cite{konevcny2016federated}.  FL allows the central machine learning model to build its learning from a broad range of data sets located at different locations. It aims to train a shared centralized machine learning model using datasets stored and distributed across multiple devices or clients. This innovative machine learning approach can train a centralized model on data generated and located on multiple clients without compromising the privacy and security of the collected data. Also, it does not require transmitting large amount of data which can be a major performance challenge especially for real-time applications. As such, FL has attracted several application domains in which data cannot be directly aggregated for training machine learning models. Furthermore, FL allows sharing learning in such cases of insufficiency of data availability and lack of other data categories where in most cases data variability is scattered among different sources. This often leads to the unsatisfactory performance of machine learning models, which has become the bottleneck of several application domains. A good example of that is in the civil infrastructures domain specifically in Structural Health Monitoring (SHM) applications where smart sensors are utilized to continuously monitor the health status of complex structures such as bridges to generate actionable insights such as damage detection. Some of these sensors may collect useful data samples that capture potential damage, where other sensors may lack for such patterns (i.e., collect only healthy data samples). In such cases, a FL approach can enable multiple sensors to collaborate on the development of a central learning model, but without needing to share or pool all the data measured from several sensors with each other. This approach can work efficiently and effectively by only sharing the model coefficients of each client model rather than the whole data collected by all participating sensors at each period of time. The learning effectiveness of the model should continue to improve over the course of several training iterations during which the shared models get exposed to a significantly wider range of data than what any single sensor node possesses locally. In this context, the effectiveness of the learning can become dependent on the distribution of the data collected by each sensor in terms of health and abnormal data points.

One of the key problems that impact data-driven machine learning approaches is identifying abnormal patterns in data. In this sense, patterns in data that do not conform to the expected normal behavior are called anomalies or outliers. The process of identifying such  non-conforming patterns is referred to as \textit{anomaly detection} \cite{chandola2009anomaly}. Anomaly detection has been useful as it  was extensively employed in a variety of applications such as intrusion detection in cyber-security and damage detection in SHM systems. One of the common challenges in anomaly detection problems is that only data representing normal behavior are available while it is difficult or even impossible to gather anomalous samples. For example, in SHM, the occurrence rate of damage is very low, and only healthy or normal data are often available. To address this challenge, a one-class learning process can be employed to construct an anomaly detection model. One-class support vector machine (OCSVM)\cite{scholkopf2001estimating,tax2004support} has been widely applied to anomaly detection and become more  popular in recent years \cite{anaissi2017adaptive,anaissi2017self,anaissi2018adaptive}.

In this context, developing an effective and efficient anomaly detection model in such applications like SHM systems requires information derived from many spatially-distributed locations throughout large infrastructure covering various points in the monitored structure. However, consolidating this data in a centralized learning model can often be computationally complex and costly. This motivates for developing a more intelligent model that utilizes the centralized learning model but without the need to transmit the frequently-measured data to one central model for processing unit.

In this study, we propose a federated learning approach to overcome the above-discussed challenges of machine learning mechanisms in centralized and distributed settings. Our approach is developed based on Federated Learning (FL) concepts for anomaly detection using OCSVM. OCSVM  has been successfully applied in many application domains such as civil engineer, biomedical, and networking \cite{khoa2017smart,anaissi2016ensemble,binbusayyis2021unsupervised}, and produced promising results. Although our approach results in reducing data transmission and improving data security, it also raises significant  challenges in how to deal with non-IID (Independent and Identically Distributed) data distribution and statistical diversity. 
The work proposed in \cite{li2019fedmd,deng2020adaptive} has thoroughly investigated the performance of the global model on clients' local data by showing how the accuracy decreases when the data diversity increases. Therefore, to address the non-IID challenge in our proposed FL approach, we developed a novel method to personalize the resulting support vectors from the FL process. The rationale idea of personalizing support vectors is to leverage the central model in optimizing the clients' models not only by using FL, but also by personalizing it w.r.t its local data distribution. We also address the communication and aggregation problems that arise in a FL approach  where several distributed models (clients) communicate with the central model to report its learning to the central model (the server). The contribution of the work in this study is threefold.

\begin{itemize}
	\item A novel method of learning OCVSM model in FL settings and an efficient communication method for aggregation.
	\item A novel method to personalize the resulting support vectors to addresses the problem of non-IID distribution of data in FL. This allows for better generalization of the clients' models and thus improves the accuracy of learning.
	
	
\end{itemize}

%% file: relatedwork.tex
\section{Related Work}
\label{s:related}

Federated Learning (FL) has gained a lot of interest in recent years and as a result, it has attracted AI researchers as a new and promising machine learning approaches \cite{kairouz2019advances,t2020personalized}. This FL approach attracts several well-suited practical problems and application areas due to its intrinsic settings where data needs to be decentralized and privacy to be preserved. However, only a few studies, that have been reported in the literature, utilized the FL approach to construct a global model. For instance, Bonawitz \etal \cite{hard2018federated} employed FL model settings to develop a system that solves the problem of next-word prediction in mobile devices. On the other hand, several other studies focused on addressing the training challenges of a central model to support all local data training especially when the distribution of data across clients is highly non-IID (independent and identically distributed).

McMahan \etal \cite{mcmahan2017communication} proposed the first FL-based algorithm named \textit{FedAvg}. It uses the local Stochastic Gradient Descent (SGD) updates to build a global model by taking average model coefficients from a subset of clients with non-IID data. This algorithm is controlled by three key parameters: $C$, the proportion of clients that are selected to perform
computation on each round; $E$, the number of training passes each client makes over its local dataset on each round; and $B$, the local mini-batch size used for the client updates. Selected
clients perform SGD locally for E epochs with mini-batch size B. Any clients which, at the start of the update round, have not completed E epochs (stragglers), will simply not be considered
during aggregation. Subsequently, Li \etal \cite{li2018federated} introduced the \textit{FedProx} algorithm, which is similar to FedAvg. However, FedProx makes two simple yet critical modifications that demonstrated performance improvements. FedProx would still consider stragglers (clients which have not completed E epochs at aggregation time) and it adds a \textit{proximal term} to the objective function to address the issue of statistical heterogeneity. Similarly, Manoj \etal \cite{arivazhagan2019federated} addressed the  effects of statistical heterogeneity problem using a \textit{personalization-based approach (FedPer)}. In their approach, a model is viewed as base besides personalization layers. The base layers will be aggregated as in the standard FL approach with any aggregation function, whereas the personalized layers will not be aggregated.  Several other methods have been proposed to achieve personalization in FL. Recently, Smith \etal \cite{smith2017federated} proposed a new algorithm named \textit{MOCHA-based multi-task learning (MTL)} framework to address the non-IID challenge in FL. Hanzel \etal \cite{hanzely2020federated} also proposed an L2GD algorithm that combines the optimization of the local and global models. Similarly, Deng \etal \cite{deng2020adaptive} developed an \textit{adaptive personalized federated learning (APFL)} algorithm which mixes the user’s local model with the global model. 

%


%% file: method.tex
\section{Personalized Conditional FedAvg: PC-FedAvg}
\label{s:method}
We propose a novel algorithm named \textit{PC-FedAvg (Personalized Conditional FedAvg)} which aims to mitigate the case where weights from low-performing clients are included in the aggregation by filtering clients based on their local performance, not stragglers. At the update round, all the clients handshake with the central model using their local loss values. Only clients that its local loss value is below the overall median loss performance will be included in the aggregation at the central model, and the rest will be dropped until the next round. The rationale behind this design is to mitigate the effects of aggregating 'bad' weights by identifying and not including them. This would cater for different local clients, with varying progress, to perform more local computation. Our PC-FedAvg approach is also centered around employing OCSVM. We address the non-IID challenge in FL settings by developing a personalized support vector algorithm  that is optimized for each distributed data model. 


\subsection{Conditional-FedAvg}
In FL setting, learning is modeled as a set of $C$ clients and a central server $S$, where each client learns based on its local data, and is connected to $S$ for  solving the following problem:

\begin{eqnarray}\label{eq:fl}
	\underset{w \in  \mathbb{R}^d }{min} f(w) := \frac{1}{C} \sum_{c=1}^{C}f_c(w_c)
\end{eqnarray}
where $f_c$ is the loss function corresponding to a client $c$ that is defined as follows: 

\begin{eqnarray}\label{eq:sgd}
	f_c(w_c) :=    \mathbb{E}[\mathcal{L}_c(w_c;x_i)]
\end{eqnarray}
where $\mathcal{L}_c(w_c;x_i)$ measures the error of the model $w_c$ (e.g. OCSVM) given the input $x_i$.
The Sequential Minimal Optimization (SMO) is often used in the support vector machine. However, in the case of the nonlinear kernel model as in OCSVM,  SMO does not suit the FL settings well. Therefore, we propose a new method for solving the  OCSVM problem in FL setting using the SGD algorithm. 

The SGD method solves the above problem defined in Equation \ref{eq:sgd} by  repeatedly updating $w$ to minimize $\mathcal{L}(w; x_i)$. It starts with some initial value of $w^{(t)}$ and then repeatedly performs the update  as follows :
\begin{eqnarray}\label{eq:sgdc}
	w^{(t+1)} :=    w^{(t)}   + \eta   \frac{\partial \mathcal{L}}{\partial w } (x_i^{(t)} ,w^{(t)} )
\end{eqnarray}

Thus, we need now  to formulate the  cost function of OCSVM defined in Equation \ref{cost} to be optimized with SGD subject to the constraints $ 0 \leq \alpha_i \leq \frac{1}{\nu n}$

\begin{eqnarray}\label{cost}
	\hspace{20pt} \max \mathcal{L}(\alpha) = \sum_{i}^n\alpha_i -  \frac{1}{2}\sum_{i}^n \sum_{j}^n \alpha_i\alpha_j K(x_i, x_j)
\end{eqnarray}

\begin{eqnarray*}
	s.t \hspace{2em} 0 \leq \alpha_i \leq \frac{1}{\nu n},\hspace{1em} \sum_{i=1}^n    \alpha_i =1,
\end{eqnarray*}
where $ K(x_i, x_j)$ is the kernel matrix and $\alpha$ are the Lagrange multipliers.

Let us assume  $\mathcal{L}(\alpha)$ is given at the Lagrange multiplier $\alpha_k$:

\begin{eqnarray*}\label{cost_k}
	\max \mathcal{L}(\alpha_k) = \alpha_k -  \frac{1}{2}\alpha_k^2 K(x_k, x_k) -  \alpha_k\sum_{i=1, i \neq k}^n \alpha_i K(x_i, x_k)
\end{eqnarray*}

The gradient of $\mathcal{L}(\alpha_k)$ at $\alpha_k$ is given as:

\begin{eqnarray}\label{grad_cost}
	\nabla \mathcal{L}(\alpha_k) = 1 - \sum_{i=1}^n \alpha_i K(x_i, x_k)
\end{eqnarray}
Starting from  an initial value of $\alpha$, the gradient descent approach successively updates  $\alpha$ as follows:

\begin{eqnarray}\label{eq:sgd_svm}
	\alpha^{t+1} =    \alpha^t   + \eta   \nabla J(\alpha^t)
\end{eqnarray}

In SGD approach,  the update rule for the k$^{th}$ component is given as:
\begin{eqnarray}\label{eq:sgdc_k}
	\alpha^{k}=    \alpha^k   + \eta   \nabla J(\alpha^k)
\end{eqnarray}

Recall that the optimization of $\alpha$ is subject to the constraints $0 \leq \alpha_i \leq \frac{1}{\nu n}$. Thus in the above update step, if $\alpha^k \le 0$  we reset it so that $\alpha^k = 0$, and if $\alpha^k \ge \frac{1}{\nu n}$ we reset it so that $\alpha^k = \frac{1}{\nu n}$. Therefore, our  OCSVM using SGD algorithm is presented in Algorithm \ref{algo:SGD}.

\begin{algorithm}[!t]
	\begin{algorithmic}
		\caption{Our OCSVM Using SGD}
		\label{algo:SGD}
		\REQUIRE: $X$, kernel function $\phi$ , $\eta$, $\epsilon$
		\STATE  $K  = \phi(x_i,x_j)_{i,j =1, \dots,n}$
		\ENSURE Initialize $\alpha$
		\STATE  $t = 0$
		\REPEAT
		\STATE  $\alpha = \alpha_t$
		\FOR{$k \leftarrow 1$ to $n$} 
		\STATE  $\alpha^{k}=    \alpha^k   + \eta  (1-\sum_{i=1}^n \alpha_i K(x_i, x_k))$
		\STATE   \textbf{if} $\alpha^k \le 0$ \textbf{then} $\alpha^k = 0$ 
		\STATE  \textbf{if} $\alpha^k \ge \frac{1}{\nu n}$ \textbf{then} $\alpha^k = \frac{1}{\nu n}$
		\ENDFOR
		\STATE  $\alpha_{t+1}= \alpha$
		\STATE  $t = t+1$
		\UNTIL $ \lVert\alpha_t - \alpha_{t+1}\rVert \leq \epsilon$ 
		
		\RETURN $\alpha$
		
	\end{algorithmic}
\end{algorithm}

In fact, the  SGD algorithm  in OCSVM focuses on optimizing the Lagrange multiplier $\alpha$ for  all patterns $x_i$ where ${x_i: i \in [n],\alpha_i > 0 } $ are called support vectors. Thus, exchanging gradient updates  in FL for averaging purposes is not applicable. Consequently, we modified the training process of SGD to share the coefficients of the features in the kernel space under the constraints  of sharing an equal number of samples across each client $C$. In this sense, our SGD training process computes the kernel matrix  $K  = \phi(x_i,x_j)_{i,j =1, \dots,n}$ before looping through the samples. Then it computes the coefficients $w$ after performing a number of epochs as follows:

\begin{eqnarray}\label{eq:wSGD}
	w^{(t+1)} =   \alpha   K ;
\end{eqnarray}
\begin{eqnarray*}
	s.t \hspace{2em}   \alpha=    \alpha   + \eta  (1-\sum_{i=1}^n w)  
\end{eqnarray*}

Each client  performs a number of $E$ epochs at each round to compute the gradient of the loss over its  local data and to send the model parameters $w^{t+1}$ to the central server $S$ along with their local loss. The server then aggregates the  gradients of the clients with a condition that a client should  have generated a loss below the overall median loss, and  applies the global model parameters update by computing the average value of all the selected clients model's parameters  as follows:
\begin{eqnarray}\label{eq:sgds}
	w^{(t+1)} :=     \frac{1}{C} \sum_{i=1}^{C} w^{(t+1)};
\end{eqnarray}
where $C$ is the number of selected clients.

The server then share the $w^{(t+1)}$ to all selected clients in which each one  performs another iteration to update $w^{(t+1)}$ but with setting  $w_i^{(t)} = w^{(t+1)}$ as defined in the traditional  FedAvg method. The full algorithm of our Conditional FedAvg learning process is given in Algorithm \ref{algo:SGD-fl}

\begin{algorithm}[!t]
	\begin{algorithmic}
		\caption{Conditional FedAvg}
		\label{algo:SGD-fl}
		
		\STATE  $K  = \phi(x_i,x_j)_{i,j =1, \dots,n}$
		\STATE \textbf{Server executes:}
		\STATE Initialize $\alpha, w$		
		\FOR{ each client $c \in C$  \textbf{in parallel}} 
		\STATE $w_c^{t+1}, loss_c^{t+1} = ClientUpdate(w^t)$
		\ENDFOR
		\STATE find the median $\mathcal{M} = median ( loss^{t+1})$
		\STATE select clients with $loss_c^{t+1}$ $\leq$  $\mathcal{M}$
		\STATE compute the average $w^{(t+1)} :=     \frac{1}{C} \sum_{c=1}^{C} w_c^{(t+1)}$
		\STATE send  $w^{(t+1)}$ to the selected clients
		\STATE
		\STATE $\textbf{ClientUpdate}(w)\textbf{:}$\\
		\REQUIRE $X $, kernel function $\phi$ , $\eta$	
		\REQUIRE $K  = \phi(x_i,x_j)_{i,j =1, \dots,n}$
		\FOR{$k \leftarrow 1$ to $n$} 
		\STATE  $\alpha^{k}=    \alpha^k   + \eta  (1-\sum_{i=1}^n w(k ,i))$
		\STATE  \textbf{if} $\alpha^k \le 0$ \textbf{then} $\alpha^k = 0$ 
		\STATE  \textbf{if} $\alpha^k \ge \frac{1}{\nu n}$ \textbf{then} $\alpha^k = \frac{1}{\nu n}$
		\ENDFOR
		\STATE 	$w =   \alpha   K $		
		\STATE  compute $loss$
		\RETURN $w$ and $loss$ to server
	\end{algorithmic}
\end{algorithm}

\subsection{Personalized Support Vectors}
Our proposed approach may work well when clients  have similar IID data.  However, it is unrealistic to assume that since data may come from different environments or contexts in FL settings, thus it can have non-IID. Therefore,  it is essential to decouple our model optimization from the global model learning in a bi-level problem depicted for personalized FL so the global model optimization is embedded within the local (personalized) models. Geometrically, the global model can be considered as a “\textit{central point}”, where all clients agree to meet, and the personalized models are the points in different directions that clients follow according to their heterogeneous data distributions. In this context, once the learning  process by the central model is converged and the support vectors are identified for each client, we perform  a personalized  step  to  optimize the support vectors on each client. Intuitively,  to generate a personalized client  model, its support vectors must reside on the surface of  the local training data  (i.e. edged support vector). Thus, we propose a new algorithm to inspect the spatial locations of the selected  support vector samples in the context of the FL settings explained above. It is intuitive that an edge support vector $x_e$ will have all or most of its neighbors located at one side of a  hyper-plane passing through $x_e$. Therefore, our edge pattern selection method constructs a tangent plane for each selected support vector ${x_i: i \in [n],\alpha_i > 0 } $ with its $k$-nearest neighbors data points. The method initially selects the  $k$-nearest data points to each support vector $x_s$, and then centralizes it around $x_s$ by computing the norm vector $v_i^n$ of the tangent plane at  $x_s$ . If all or most of  the vectors  are located at one side of the  tangent plane), we consider $x_s$  as an edge  support vector denoted by $x_e$, otherwise, it is considered as an interior support vector and it is excluded from the selected original set of support vectors. 



The  algorithm of selecting the edge samples is described as follows: given a set of support vectors  $x_i(i = 1, \ldots, n)$, the norm vector $v_i^n$ of the tangent plane passing through $x_i$  is computed using its  $k$-nearest neighbors $(x_{ij}, i = j=1,\ldots,k)$ as follows:

\begin{eqnarray}\label{norm1}
	v_i^n = \sum_{j=1}^{k}v_{ij}^u
\end{eqnarray}
where 
\begin{eqnarray}\label{norm2}
	v_{ij}^u = (x_{ij}-x_i)/ \Vert x_{ij}-x_i \Vert  
\end{eqnarray}

Then we calculate $\theta_{ij}$ the dot products between the normal vector $v_i^n$  and the vectors from $x_i$ to its $k$-nearest neighbors as follows:

\begin{eqnarray}\label{norm3}
	\theta_{ij} = (	v_{ij}^u )^T   v_i^n
\end{eqnarray}

If the sign of the product $\theta_{ij}$ between $v_{ij}^u$ and $v_i^n$ is greater than zero, then pattern $x_{ij}$ is on the positive side of the tangent plane. If $\theta_{ij} = 0$, it is on the tangent plane. Thus, an edge pattern can be determined by counting the number of neighbors with $\theta_{ij} \geq 0$  $(0<\theta_{ij}<\pi/2)$ as follows:

\begin{eqnarray}\label{rate}
	l_i =     \frac{1}{k}\sum_{j=1}^{k} I ( \theta_{ij} ) \geq  0
\end{eqnarray}
where $I(.)$ is an indicator function incremented by one when the condition is true.
If $x_i$ is located at a  convex surface, then  all  the $k$-nearest neighbors of $x_i$  should be  located at one side of the tangent plane, then we count  $x_i$ as an edge support vector. If $x_i$ is located at a  concave  surface,  it is  also  considered as  an edge support vector but with a small ratio of the $k$-nearest neighbors located at the other side of the tangent plane.  Therefore, to identify  an edge support vector in concave case, we  use a threshold $1-\gamma$ ($\gamma$ is a small positive parameter) to represent the  acceptable ratio of samples located at the other side of the tangent plane  for each sample $x_i$.  For $\gamma =0.05$, if 95\% of the $k$-nearest neighbours to a sample $x_s$ are  are located at one side, then the sample $x_i$ can still be considered as an edge support vector, otherwise $x_s$ is considered as an interior support vector.



\begin{algorithm}[!t]
	\begin{algorithmic}
		\caption{Our Personalized Support Vectors Algorithm}
		\label{es}
		\REQUIRE Get the set of support vectors  $\{{x_{i}}\}_{i=1}^{n_s}$.
		\FOR{ $x_{i}(i = 1, \ldots, n_{s})$}		
		\STATE Find the $k$ closest points to $x_{s}$: $x_{j}, j= 1, \dots,k$.
		\STATE Calculate the normal vectors $v_i^n$ of  $x_{s}$ according to (\ref{norm1}).
		\STATE calculate the dot products $\theta_{ij}$ according to \ref{norm3}https://www.overleaf.com/project/6152a9adf1c6ae66769c8da8
		\STATE counting the number of neighbors with $\theta_{ij} \geq 0$ according to (\ref{rate}).
		\IF { $l_i \leq 1 - \gamma$ } 
		\STATE $x_{i}$ is an edge support vector
		\ELSE
		\STATE   $x_{i}$ is identified as non-support vector.
		\ENDIF
		\ENDFOR	
		\RETURN  $\{{x_{s}}\}_{s=1}^{n_s^{new}}$
	\end{algorithmic}
\end{algorithm}

%% file: experiments.tex
\section{Experimental Results and Discussions}
\label{s:results}

We validate our PC-FedAvg method using two  real SHM datasets collected from structural bridges to detect potential damage in these bridges.  

\subsection{Experiments on Real SHM Data}

Besides the experiment on the toys dataset, we validate our PC-FedAvg method using real SHM data collected  from Bridge structure. The main goal is to detect potential damage in bridge structures. We conduct experiments on two case studies using structural vibration based datasets acquired from a network of accelerometers mounted on two  bridges in Australia, Cable-Stayed Bridge and Arch Bridge\footnote{The two bridges are operational and the companies which monitor them requested to keep the bridge name and the collected data about its health confidential.}. In all experiments, we used the default value of the Gaussian  kernel parameter $\sigma$  and $\nu = 0.05$. The accuracy values were obtained using the F-Score (FS),  defined as $\textrm{\Fscore} = 2 \cdot \dfrac{\textrm{Precision}  \times \textrm{Recall} }{\textrm{Precision} + \textrm{Recall}}$ where $\textrm{Precision} = \dfrac{\textrm{TP} }{\textrm{TP} + \textrm{FP}}$ and $\textrm{Recall}  = \dfrac{\textrm{TP} }{\textrm{TP} + \textrm{FN}}$ (the number of true positive, false positive and false negative are abbreviated by TP, FP and FN, respectively). 

\subsubsection{\textbf{Experimental Results on Cable-Stayed Bridge}}
\label{s:data_wsu}

\begin{figure*}
	\centering
	\includegraphics[scale=0.3]{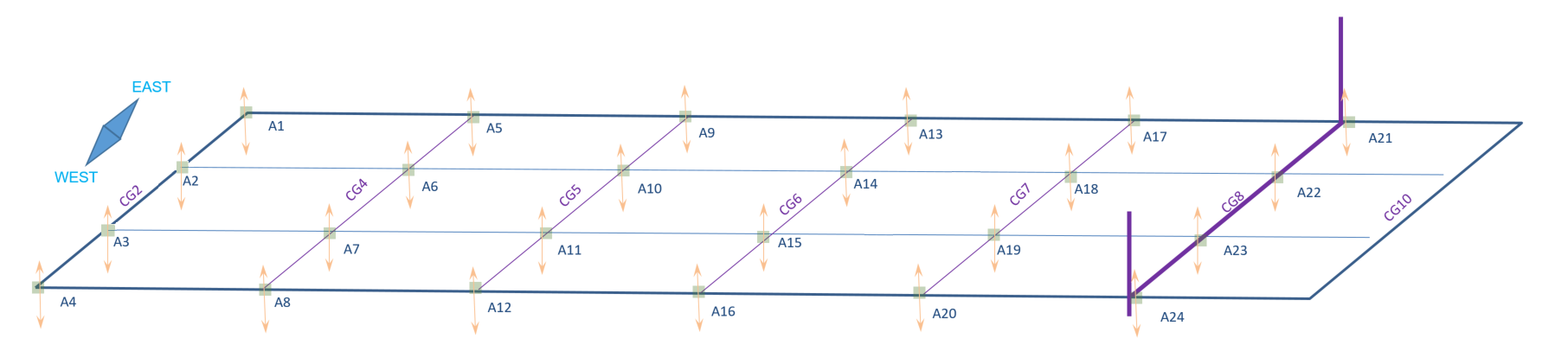}
	\caption{The locations on the bridge's deck of the 24 $Ai$ accelerometers used in this study. The cross girder $j$ of the bridge is displayed as $CGj$ \cite{anaissi2018tensor}.}
	\label{fig:wsuloc}
\end{figure*}

We instrumented the Cable-Stayed Bridge with 24 uni-axial accelerometers and 28 strain gauges. We used accelerations data collected from sensors $Ai$ with $i\in [1;24]$.   Figure~\ref{fig:wsuloc} shows the locations of these 24 sensors on the bridge deck. Each set of sensors on the bridge along with one line (e.g A1: A4) is connected to one client node and fused in a tensor node  $\mathcal{T}$ to represent one client in our FL network, which results in six tensor nodes  $\mathcal{T}$ (clients).

In this bridge dataset, two different types of damage were emulated (for experimentation purposes) by placing a large static load (a vehicle) at different locations of the bridge. Three scenarios were emulated namely: no vehicle is placed on the bridge (\textit{"healthy state"}), a light vehicle with the approximate mass of 3 ton is placed on the bridge close to location A10 (\textit{"Car-Damage"}) and a bus with the approximate mass of 12.5 t is located on the bridge at location A14 (\textit{"Bus-Damage"}).  This emulates slight and severe damage cases which were used in our performance evaluation.   

This experiment generates 262 samples (a.k.a events) each of which consists of acceleration data for 2 seconds at a sampling rate of 600 Hz. We separated the 262 data instances into two groups, 125 samples related to the healthy state and 137 samples for the damage state. The 137 damage examples were further divided into two different damaged cases: the "Car-Damage" samples (107)   generated when a stationary car was placed on the bridge, and the "Bus-Damage" samples (30) emulated by the stationary bus. 

For each reading of the uni-axial accelerometer, we normalized its magnitude to have a zero mean and one standard deviation. The fast Fourier transform (FFT) is then used to represent the generated data in the frequency domain.  Each event now has a feature vector of  600 attributes representing its frequencies. The resultant data at each sensor node $\mathcal{T}$ has a structure of 4 sensors $\times$ 600 features  $\times$ 262 events.

We separated the 262 data instances into two groups,   125 samples related to the healthy state and 137 samples for the damage state. The 137 damage examples were further divided into two different damaged cases: the "Car-Damage" samples (107)   generated when a stationary car was placed on the bridge, and the "Bus-Damage" samples (30) emulated by the stationary bus. 

We randomly selected 80\% of the healthy events (100 samples) from each tensor node $\mathcal{T}$ for  training multi-way of $ \mathcal{X} \in \mathbb{R}^{4 \times 600 \times 100}$ (i.e. \textit{training} set). The 137 examples related to the two damage cases were added to the remaining 20\% of the healthy data to form a \textit{testing} set, which was later used for the model evaluation. 

At each client node $\mathcal{T}$, we implemented our Algorithm \ref{algo:SGD-fl}  to construct an OCSVM  model at each client as well as the central model followed by  Algorithm \ref{es} to personalize the obtained support vectors to each client's data.

\begin{figure*}[!t]
	\centering
	\captionsetup[subfloat]{}
	\subfloat[ The effect of number of commuication rounds.]{{\includegraphics[width=2.5in,height=1.6in]{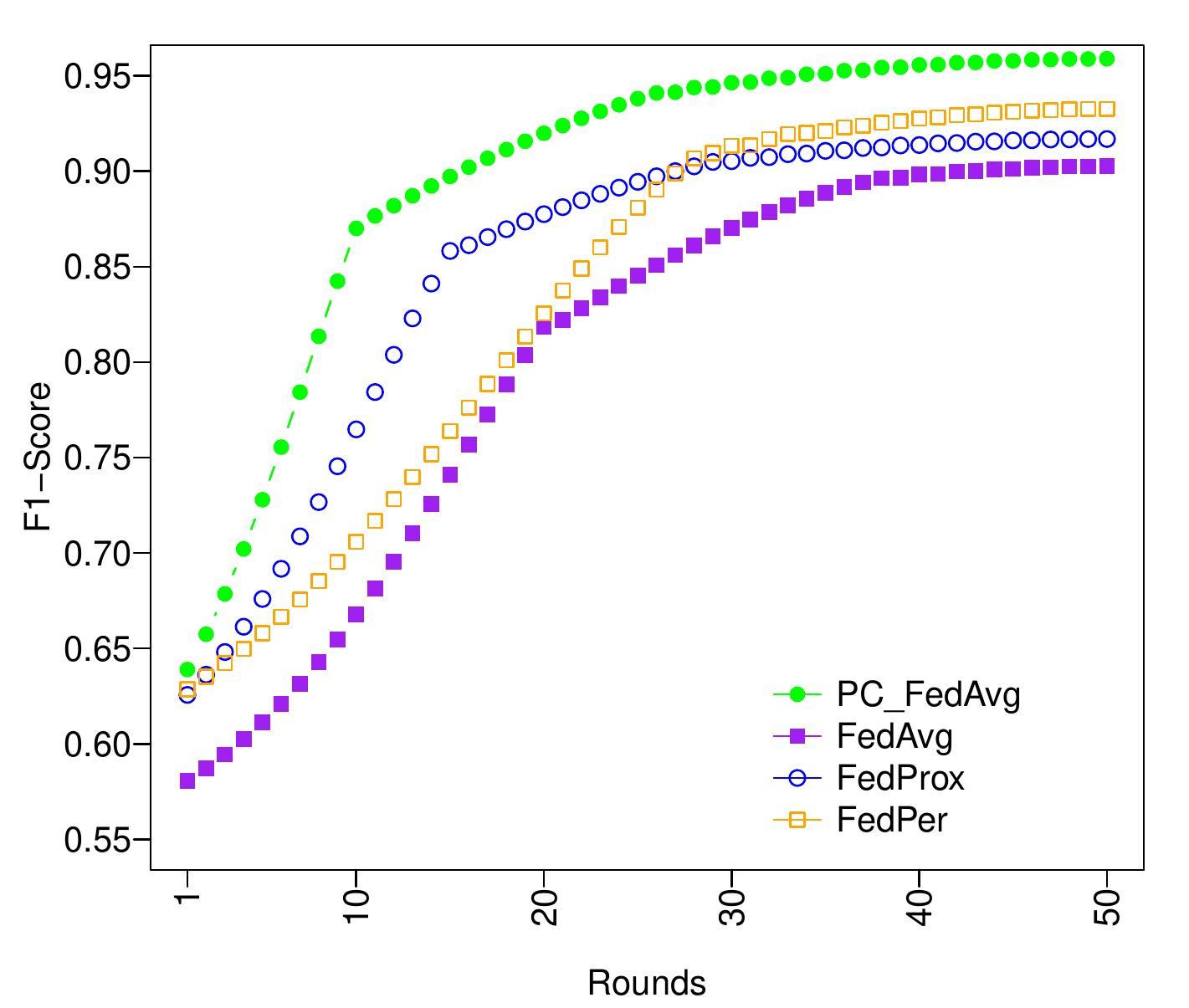} }}%
	\subfloat[The effect of number of local training epochs.]{{\includegraphics[width=2.5in,height=1.6in]{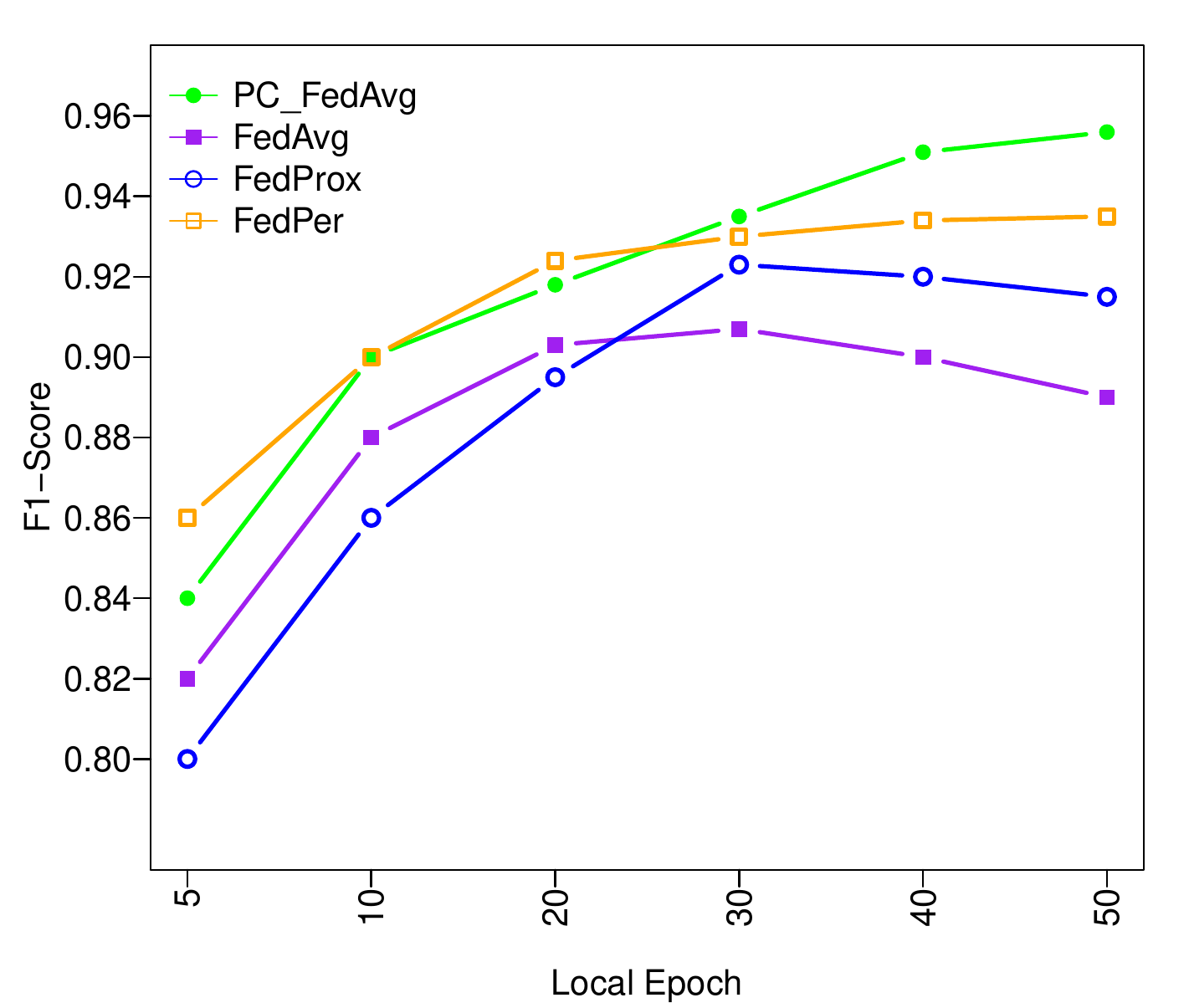} }}%
	\caption{ Convergence rates of various methods in  federated learning applied on Cable-Stayed Bridge with $\mathcal{T}$ = 6 clients.}%
	\label{conv_cable}%
\end{figure*}

We initially study the effect of the  number of local training epochs $E$ on the  performance of the four experimented federated learning  methods as suggested in previous works \cite{mcmahan2017communication,chen2018federated}. The candidate local epochs we consider are  $E \in {5, 10, 20,30, 40, 50}$. For each of the candidate $E$, we run all the methods  for 40  rounds  and report the final f1-score accuracy generated by each method. The result is shown in Figure \ref{conv_cable}(b).  We observe that conducting  longer epochs on the clients improves the performance of PC-FedAvg and FedPer, but it slightly deteriorates the performance of FedProx and  FedAvg. The second experiment was  to compare our method to FedAvg, FedPer and FedProx in terms of accuracy  and the number of communication rounds  needed for the global model to achieve good performance on the test data. We set the total number of epochs $E$  for PC-FedAvg  and FedPer to 50, and  30  for FedProx and FedAvg as determined by the  first experimental study related to the  local training epochs $E$.  The results showed that PC-FedAvg outperforms FedAvg,  FedProx and FedPer in terms of local training models  and performance  accuracy. Table \ref{results} shows the accuracy results of all experiments using $\Fscore$.  Although no data from the damaged state has been employed to construct the central model, each personalized local client model was able to identify the damage events related to "Car-Damage"  and "Bus-Damage"   with an average $\Fscore$ accuracy of $0.96\pm0.02$.

\subsubsection{\textbf{Experimental Results of The Arch Bridge:}}
\label{s:data_arch}

\begin{figure*}[!t]
	\centering
	
	\includegraphics[width=4.5in]{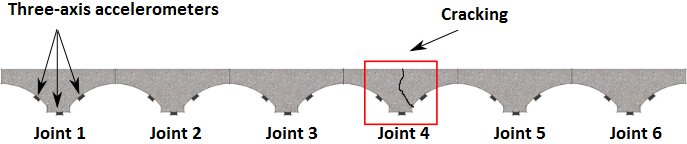}
	
	\caption{Evaluated joints on the Arch Bridge.}
	
	\label{bridge}
\end{figure*}

The Arch bridge has 800 jack arches to support its joints. Each joint is instrumented by three tri-axial accelerometers mounted on the left, middle and right side of the joint, as shown in Fig \ref{bridge}. We conducted two different experiments using this data. This experiment uses six joints (named 1 to 6) where only one joint (number four) was known as a cracked joint. The data used in this study contains  36952 events as shown in Table \ref{bridge_data} which were collected for three months. Each event is recorded by a sensor node when a vehicle passes by a jack arch for 1.6 seconds at a sampling rate of 375 Hz resulting in a feature vector of  600 attributes in the time domain.  All the events in the datasets (1, 2, 3, 5, and 6) are labelled positive (healthy events), where all the events in dataset 4 (joint 4) are labelled negative (damaged events). For each reading of the tri-axial accelerometer (x,y,z), we calculated the magnitude of the three vectors and then the data of each event is normalized to have a zero mean and one standard deviation. Since the accelerometer data is represented in the time domain, it is noteworthy to represent the generated data in the frequency domain using Fourier transform. The resultant six datasets (using the middle sensor of each joint) have 300 features that represent the frequencies of each event.

\begin{table}
	\centering
	\caption{Number of samples at each joint.}
	\label{bridge_data}
	\begin{footnotesize}
		\begin{tabular}{c | c c c} \hline
			Dataset & Number of samples & Training & Test\\ \hline
			Joint 1 &  6329             & 4430     & 1899 \\
			Joint 2 &  7237             & 5065     & 2172\\
			Joint 3 &  4984             & 3488     & 1496\\
			Joint 4 &  6886             & 0        & 6886 \\
			Joint 5 &  6715             & 4700     & 2015\\
			Joint 6 &  4801             & 3360     & 1441\\
			
			\hline\end{tabular}
	\end{footnotesize}
\end{table} 

For each dataset, we randomly selected 80\% of the positive events for training and 20\% for testing in addition to the unhealthy events in dataset 4.  Like the Toys dataset experiment, we applied our  Algorithms \ref{algo:SGD-fl} to learn a federated OCSVM model at each client’s model and at the server’s model. We also applied our Algorithm \ref{es} to personalize each client's support vectors.  

\begin{figure*}[!t]
	\centering
	\captionsetup[subfloat]{}
	\subfloat[ The effect of number of commuication rounds.]{{\includegraphics[width=2.5in,height=1.6in]{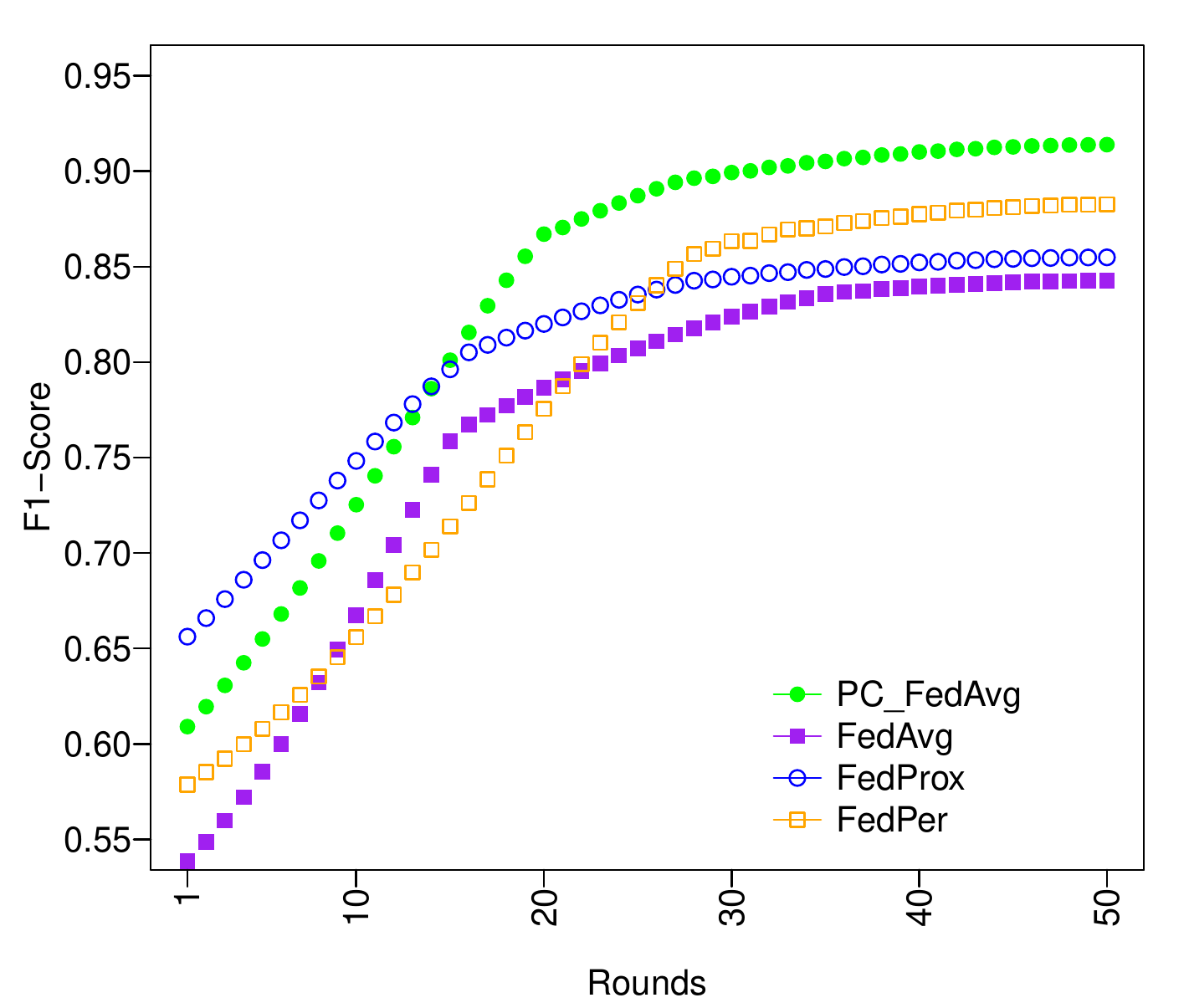} }}%
	\subfloat[The effect of number of local training epochs.]{{\includegraphics[width=2.5in,height=1.6in]{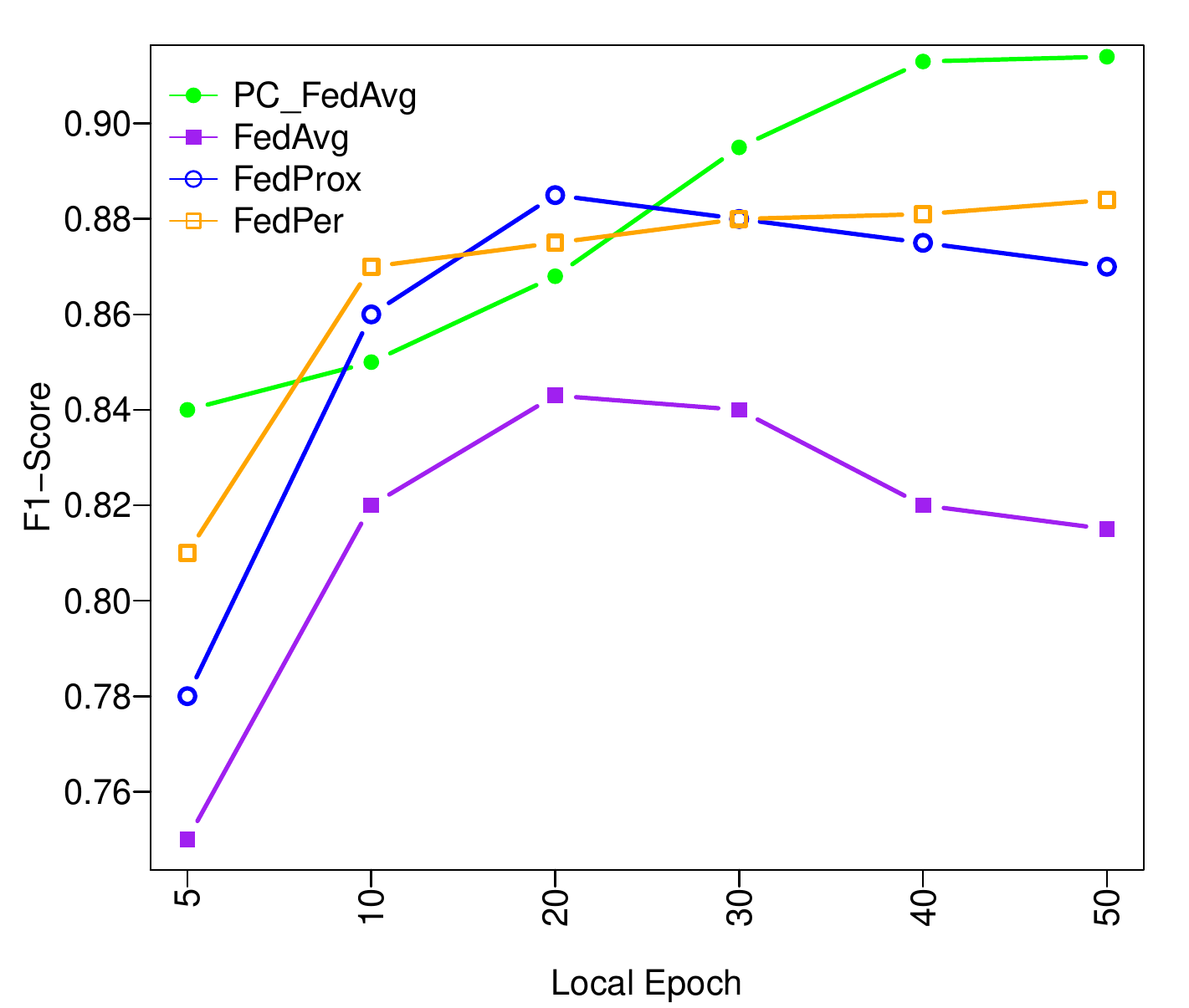} }}%
	\caption{ Convergence rates of various methods in  federated learning applied on The Arch Bridge with $\mathcal{T}$ = 6 clients.}%
	\label{conv_arch}%
\end{figure*}

Similar to the previous dataset, we initially study the effect of the  number of local training epochs $E$ on the  performance of the four experimented FL  methods. The results are summerised in Figure \ref{conv_cable}(b). Both PC-FedAvg  and FedPer benefit from doing   longer epochs on clients in  contrast to   FedProx and  FedAvg  which are slightly  deviated from the optimal convergence.  The second experiment was  to compare our method to FedAvg, FedPer and FedProx in terms of accuracy  and the number of communication rounds  needed for the global model to achieve good performance on the test data. We set the total number of epochs $E$  for PC-FedAvg  and FedPer to 40, and  20  for FedProx and FedAvg as suggested from Figure \ref{conv_arch}(b). As shown, PC-FedAvg outperforms FedAvg,  FedProx and FedPer in terms of local training models  and performance  accuracy. Table \ref{results} shows the accuracy results of all experiments using $\Fscore$.

We observed that each personalized client's model was able to identify its local healthy samples with an average $\Fscore$  equal to $0.91\pm0.03$. Furthermore, the model at the client that represents joint 4 was also able to identify $0.90$ of the damage samples. These results demonstrate that the PC-FedAvg approach without data sharing is still able to identify damage events even though these events were not involved in the training process. The results also clearly show that learning from massively decentralized data is still challenging and needs improvement especially in the prediction accuracy of damage events at joint 4. Table \ref{results} shows the resulting  FP, TP, and $\Fscore$ accuracies of the PC-FedAvg method compared to FedAvg,  FedProx and FedPer methods. It can be concluded from this table that PC-FedAvg outperformed other state-of-the-art methods in detecting  damage in the bridge structure. This is evident by the higher accuracy score and the lower false alarm rates of our PC-FedAvg. \\

As validated by the two experiment sets, our PC-FedAvg outperforms alike FL methods in terms of accuracy and personalization of learning with one OCSVM. Like traditional FL approaches, our model reduces frequent data transmission and preserving the privacy of the data as learning occurs at the clients’ models and the learning co-officiants are shared. On the other hand, our model has a unique advantage compared to the other state-of-the-art approaches. Specifically, the personalization effect is fundamental toward achieving better accuracy as our PC-FedAvg personalizes the resultant support vectors to address the problem of non-IID distribution of data in normal FL, and thus it achieves better generalization of the clients' models. In practice, data collected at different clients (e.g., devices or sensors) is likely to exhibit non-IDD and be statistically diverse. This would impact the accuracy of non-personalized FL methods as it is evident that the accuracy of a global model based on clients' local data would decrease when the data diversity increased \cite{li2019fedmd,deng2020adaptive}. Our  approach is centered around employing OCSVM as anomaly detection model for damage detection leads to many benefits from the computational perspective, the low resources needed to run the OCSVM algorithm lets to participate in the collaborative learning even though meager resources devices, especially when the type of the kernel function used by the machine learning  algorithm is linear. Moreover, the  approach we follow of  mitigating  the effects of aggregating 'bad' weights by identifying and not including clients   based on their local performance  which  allow us more freedom to perform more local computation. Therefore, our PC-FedAvg is essential to deal with such data issues and to improve the learning effect in FL approaches.

\begin{table}[!t]
	\centering
	\caption{ $\Fscore$ of various methods. }
	\label{results}
	\renewcommand{\arraystretch}{1.5}
	\begin{small}
		\begin{tabular}{| m{0.7cm}|  m{1.6cm}| m{1.2cm} | m{1.2cm} | m{1.2cm} |}
			\hline
			&PC-FedAvg & FedProx & FedPer & FedAvg   \\
			\hline
			
			Cable & 0.96$\pm$0.02  & 0.92$\pm$0.01 & 0.93$\pm$0.03 & 0.90$\pm$0.04      \\

			\hline
			
			Arch  &  0.91$\pm$0.03 &  0.85$\pm$0.02 & 0.88$\pm$0.05  & 0.84$\pm$0.06 \\ 
			
			\hline
		\end{tabular}
	\end{small}
\end{table}

%% file: conclusion.tex
\section{Conclusions}
\label{s:conclusion}

In this paper, we present a novel machine learning approach for an effective and efficient anomaly  detection model in such applications like SHM systems that require information derived from many spatially-distributed locations throughout large infrastructure covering various points in the monitored structure. Our method employs a Federated Learning (FL) approach to OCSVM as an anomaly detection model augmented with a method to personalize the resulting support vectors from the FL process.  Our experimental evaluation on two real bridge structure datasets showed promising damage detection accuracy by considering different damage scenarios. In the "Cable-Stayed Bridge" dataset, our PC-FedAvg method achieved an accuracy of  96\%.  In the Arch Bridge dataset, our PC-FedAvg method achieved 91\% damage detection accuracy. The experimental results of these case studies demonstrated the capability of our FL-based damage detection approach with the personalization algorithm to improve the damage detection accuracy. 